\documentclass[letterpaper]{article} 
\usepackage{aaai2026}  
\usepackage{times}  
\usepackage{helvet}  
\usepackage{courier}  
\usepackage[hyphens]{url}  
\usepackage{graphicx} 
\usepackage{natbib}  
\usepackage{amsmath}
\usepackage{caption} 
\usepackage{amssymb}
\usepackage{xspace}
\usepackage{booktabs}
\usepackage{algorithm}
\usepackage{algorithmic}

\usepackage{newfloat}
\usepackage{listings}
\DeclareCaptionStyle{ruled}{labelfont=normalfont,labelsep=colon,strut=off} 
\floatstyle{ruled}
\newfloat{listing}{tb}{lst}{}
\floatname{listing}{Listing}
\title{Generating Surface for Text-to-3D using 2D Gaussian Splatting}
\author{
    Huanning Dong\textsuperscript{\rm 1}, Fan Li\textsuperscript{\rm 1}, Ping Kuang\textsuperscript{\rm 1}, Jianwen Min\textsuperscript{\rm 1}
}
\affiliations{
    \textsuperscript{\rm 1}University of Electronic Science and Technology of China\\

}

\usepackage{bibentry}

\begin{document}

\maketitle

\begin{abstract}
Recent advancements in Text-to-3D modeling have shown significant potential for the creation of 3D content. However, due to the complex geometric shapes of objects in the natural world, generating 3D content remains a challenging task. Current methods either leverage 2D diffusion priors to recover 3D geometry, or train the model directly based on specific 3D representations. In this paper, we propose a novel method named DirectGaussian, which focuses on generating the surfaces of 3D objects represented by surfels. In DirectGaussian, we utilize conditional text generation models and the surface of a 3D object is rendered by 2D Gaussian splatting with multi-view normal and texture priors. For multi-view geometric consistency problems, DirectGaussian incorporates curvature constraints on the generated surface during optimization process. Through extensive experiments, we demonstrate that our framework is capable of achieving diverse and high-fidelity 3D content creation.
\end{abstract}
\begin{figure*}
\centering
\includegraphics[width=1\linewidth]{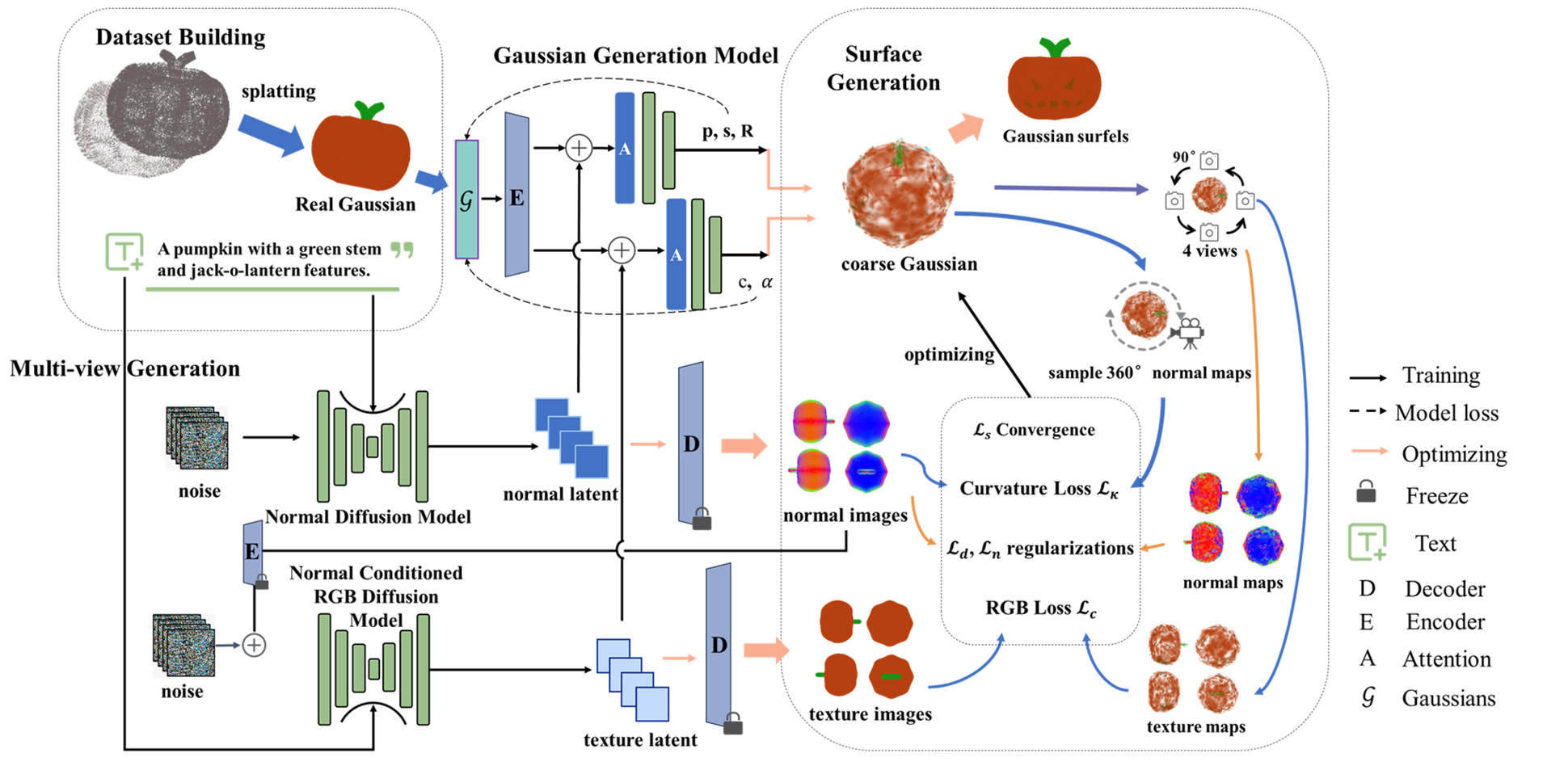}
\caption{DirectGaussian generates 3D objects from a natural language caption such as ``A pumpkin with a green stem and jack-o'-lantern features”. The coarse Gaussian is produced by a multi-head attention model, whose hidden representations incorporate normal and texture latents extracted from two multi-view diffusion models, both conditionally guided by text. DirectGaussian refines the coarse Gaussians by minimizing splatting loss, as well as surface convergence and curvature constraints that enforce geometric consistency across 360-degree perspectives.}
\label{fig:Framework}
\end{figure*}
\section{Introduction}
\label{sec:intro}

Text-to-3D generation has emerged as a challenging task in generative artificial intelligence. Direct approaches typically involve training 3D diffusion models on 3D datasets \cite{shapenet,objaverse}. Many studies have focused on generating point clouds directly from text, leveraging the relative ease of acquiring point cloud data compared to other 3D formats. However, this often results in discrete and unrealistic visual outputs.

To address this, various alternative 3D representations have been explored for object generation, including neural radiance fields (NeRF), voxel grids, and polygon meshes. Nevertheless, NeRF suffers from high rendering costs, while mesh-based methods are limited by their reliance on predefined topologies, which hinders large-scale practical deployment. As a more efficient 3D representation, Gaussian Splatting has recently emerged as a promising solution for 3D generation tasks. Prior work \cite{high,Surfacesplatting,2dsplatting} has demonstrated that surfels (surface elements \cite{surfels,phong}) can serve as an effective representation of complex geometries.

Inspired by the original rendering mechanism of Gaussian Splatting, reconstructing 3D objects from generated multi-view images is a natural choice. Many recent works follow this paradigm by generating multi-view images from text. However, the limited number of viewpoints significantly degrades Gaussian rendering quality (e.g., insufficient texture detail and noticeable noise when viewed from novel angles). Increasing the number of text-generated views is infeasible, as doing so makes maintaining geometric consistency across all views difficult. Furthermore, due to the inherent complexity of 3D model structures and the scarcity of large-scale 3D datasets, directly obtaining reliable 3D representations from text remains challenging.

In our approach, we parameterize 3D object surfaces using Gaussian surfels \cite{surfels,phong}. We fine-tune a pretrained multi-view normal and texture diffusion model to provide 2D priors, which are jointly trained with our proposed Gaussian surfel generation model. Our key motivation is to generate Gaussian surfels that encode multi-view 2D information in a geometrically consistent manner, thereby producing high-quality initial Gaussian inputs for the splatting process. This is critical, as splatting not only depends on rendering outcomes but also requires robust initial values that strongly influence geometric and textural fidelity. For instance, the original Gaussian Splatting method uses Structure-from-Motion (SfM) \cite{sfm} to reconstruct an initial point cloud. In our case, the use of multi-view normal and texture priors enables us to render initial Gaussians aligned with corresponding image conditions, facilitating parameter optimization.

Moreover, to avoid high calculation costs and time, we construct a Gaussian surfel dataset of objects conforming to textual descriptions. Given our four-view rendering strategy, the original splatting loss proves insufficient. To address this, we introduce 360° surround-view surface curvature constraints and surface convergence constraints, which help enhance the geometric and textural quality of the generated objects.

Our main contributions are summarized as follows:

1. We propose an efficient text-to-3D generation framework, where coarse Gaussians obtain basic geometric and textural priors derived from a curated surface parameter dataset and align with the input text.

2. By rendering the coarse Gaussians to match multiview normal and texture images, we introduce a 360° surround-view surface curvature loss that enforces curvature-based normal consistency. This loss, together with surface convergence constraints, helps preserve fine geometric details in the final outputs.

3. Our method, combined with 2D Gaussian Splatting, is extensively evaluated under various text prompts. The resulting system, DirectGaussian, demonstrates new potential for differentiable surface generation in text-to-3D tasks.

\begin{figure*}
\centering
\includegraphics[width=0.9\linewidth]{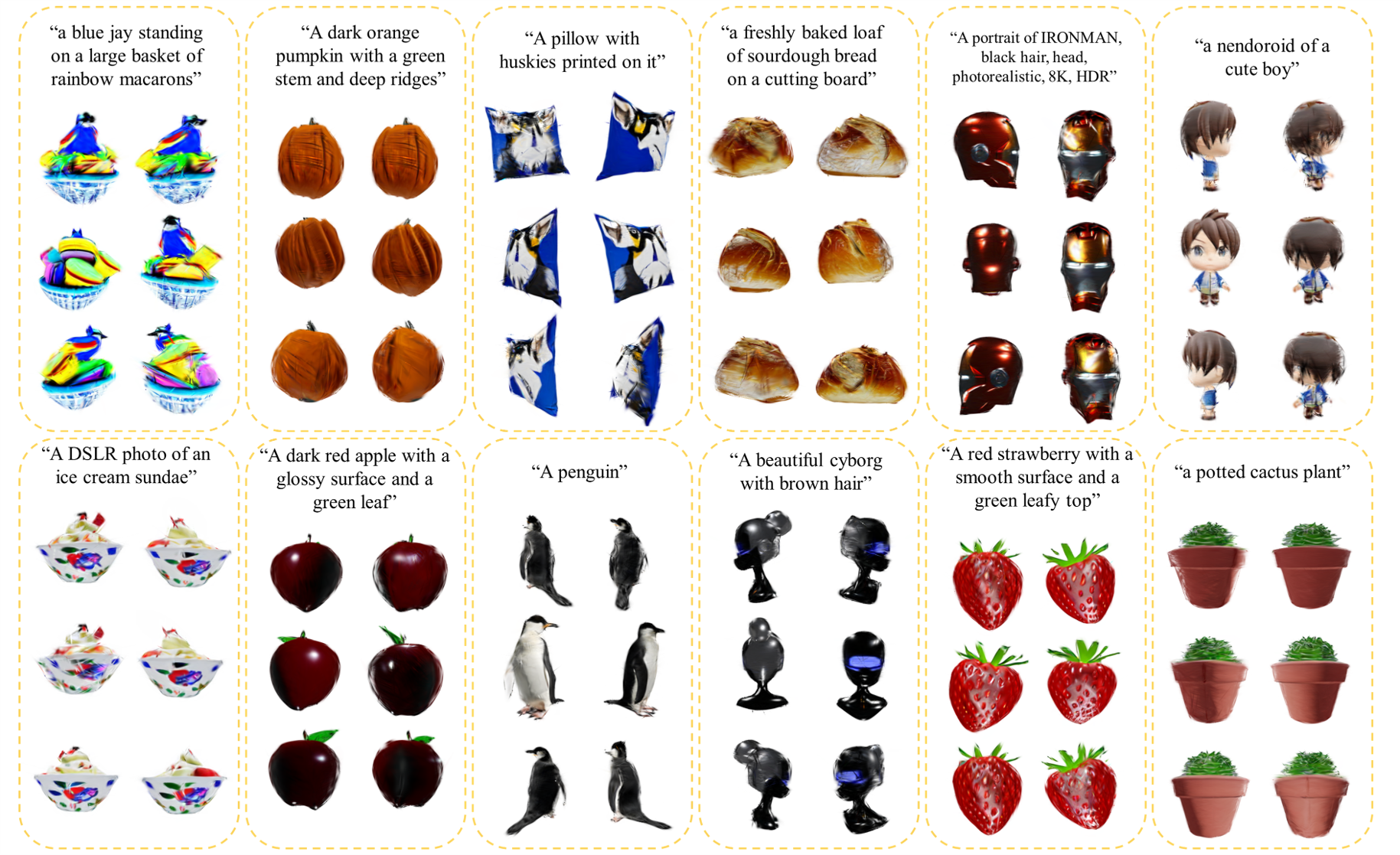}
\caption{The gallery of text-to-3d generation from DirectGaussian. Given text prompts as description input, our method outputs high-quality textured 3D objects within minutes. The text prompt is selected from historical text-to-3D works.}
\label{fig:results}
\end{figure*}

\section{Related Work}
\label{sec:RelatedWork}
\subsection{Gaussian Splatting}

3D Gaussian Splatting (3DGS) \cite{3DGaussian} achieves high-fidelity novel view rendering and fast training, which has attracted widespread attention. Despite its impressive results, 3DGS requires an additional input of an initial point cloud, unlike NeRF \cite{nerf}. In the presence of noise or when using a randomly initialized point cloud, 3DGS often suffers a significant drop in performance. RAIN-GS \cite{rain}, which focuses on guiding Gaussians to learn in a coarse-to-fine manner, reveals that SfM \cite{sfm} initialization provides low-frequency components of the true distribution. However, the original 3DGS optimization scheme struggles to move Gaussians far from their initialized positions, leading to degraded performance when starting from random or noisy SfM initializations.

Furthermore, 3D Gaussians struggle to capture fine-grained geometry, as volumetric Gaussians conflict with the thin and detailed nature of surfaces. Some works have therefore explored surface reconstruction based on 3DGS. PGSR \cite{pgsr} compresses 3D Gaussians into flat planes and introduces a multi-view regularization technique to preserve global geometric accuracy. 2D Gaussian Splatting (2DGS) \cite{2dsplatting} represents a 3D scene using 2D Gaussian primitives, where each defines an oriented elliptical disk. Through the use of depth distortion and normal consistency techniques, 2DGS achieves smoother surface reconstructions via gradient-based optimization, demonstrating that 2D Gaussian primitives offer more accurate geometric representations during rendering.

\subsection{2D Images to 3D}

Zero-1-to-3 \cite{zero123} suggests that large diffusion models, though trained only on 2D images, have implicitly learned rich 3D priors about the visual world. It manipulates the camera viewpoint to perform single-image-to-3D conversion via zero-shot novel view synthesis. Building on this, One-2-3-45 \cite{one2345} leverages multi-view 3D reconstruction techniques to generate a 3D mesh from the multi-view predictions produced by Zero-1-to-3. In contrast, Zero123++ \cite{zero123++} revisits Zero-1-to-3 and fine-tunes a new multi-view consistent diffusion model based on Stable Diffusion \cite{stablediffusion}.

However, most methods that reconstruct geometry from multi-view images suffer from the so-called multi-face Janus problem \cite{prolificdreamer,magic3d,wang2023score,fantasia3d}, caused by the absence of explicit 3D supervision. To address this, Direct2.5 \cite{direct2} explores the 2.5D domain by injecting multi-view information into the self-attention layers, aligning texture maps with corresponding normal maps. Motivated by this, we adopt a joint multi-view distribution of normal and image maps to achieve geometrically consistent 3D generation.

\subsection{Text to 3D}

Based on a pretrained text-to-image diffusion model, DreamFusion \cite{dreamfusion} proposes the Score Distillation Sampling (SDS) technique to optimize a NeRF for text-to-3D synthesis. Most recent methods that produce high-fidelity 3D models rely on SDS-based optimization strategies.

DreamGaussian \cite{dreamgaussian} takes a different route by producing coarse 3D Gaussians via an image-to-3D pipeline and designing an efficient mesh extraction algorithm using local density queries. In contrast, LGM \cite{lgm} trains a network with an asymmetric U-Net backbone on multi-view image datasets to predict 3D Gaussian features. Combining 3D and 2D diffusion via Gaussian Splatting, GaussianDreamer \cite{gaussiandreamer} bridges 2D and 3D diffusion models, while Gsgen enriches coarse geometry details through a two-stage optimization process. Avoiding mesh optimization altogether, LucidDreamer \cite{luciddreamer} reveals that the core mechanism of SDS is to match rendered images with pseudo-ground-truth and proposes a novel training strategy called Interval Score Matching (ISM).

Following the multi-view generation strategy of Direct2.5 \cite{direct2}, we use normal and image maps as pseudo-ground-truth to generate surfel parameters. Built upon 2DGS, our pipeline avoids SDS optimization yet still reconstructs geometrically complex and diverse object surfaces.
\section{Method}
\label{sec:Method}
In this section, we introduce our Text-to-3D surface generation framework. As shown in Figure~\ref{fig:Framework}, DirectGaussian consists of three main components: (1) constructing a Gaussian surfel dataset aligned with textual descriptions, (2) a coarse Gaussian surfel generative model, and (3) an optimization process for the coarse Gaussians guided by multi-view priors using splatting loss and surface constraints.

The Gaussian surfel dataset is detailed in the experimental section. We begin by revisiting the surface model, which defines the surfel parameters that need to be generated and optimized. Then we introduces our generative model and our optimization process using 2D Gaussian Splatting method. 

\subsection{Surfel representation of 3D object} 
\label{sec:Surfel}
The surface of a 3D object is represented as a parameterized surface embedded in three-dimensional Euclidean space, denoted as \( S \hookrightarrow \mathbb{E}^3 \). For an arbitrary point \( \mathbf{p} \in \mathbb{R}^3 \) on \( S \), \( (u,v) \) denotes the local coordinate of the point \( \mathbf{p} \), where \( \mathbf{p} \) has two orthogonal principal tangent directions \( \mathbf{t}_u \) and \( \mathbf{t}_v \), with \( \mathbf{t}_u \times \mathbf{t}_v \neq 0 \). Following the notation in~\cite{phong}, a \emph{surfel} is defined by its center \( \mathbf{p} \) in the local frame, which is spanned by the tangent directions \( \mathbf{t}_u \) and \( \mathbf{t}_v \).

For an arbitrary point \( \mathbf{q} \) in the interior of the surfel, the local coordinates \( (u, v) \) correspond to the orthogonal projection of \( \mathbf{q} \) onto the tangent plane. The constraint \( u^2 + v^2 \leq 1 \) holds within the surfel domain. In this case, if \( s_1 \) and \( s_2 \) are two scalar values, then a surfel \( TpG \) centered at \( \mathbf{p} \) is defined as:
\begin{equation}
\label{eq:surfel}
TpG: \mathbf{p} + s_1 \mathbf{t}_u u + s_2 \mathbf{t}_v v.
\end{equation}

In this local frame, the parameterized surface model can thus be represented by a rotation matrix \( \mathbf{R} = [\mathbf{t}_u\ \mathbf{t}_v\ \mathbf{n}_p] \) and a scaling vector \( \mathbf{s} = (s_1, s_2) \). The tuple \( \{\mathbf{p}, \mathbf{s}, \mathbf{R}\} \) comprises learnable parameters that describe the geometry and structural content of the object. In analogy to 3D Gaussian Splatting~\cite{3DGaussian}, the learnable opacity and color parameters \( \{\alpha, c\} \) are used to represent the textural content of the object.

\subsection{Coarse Gaussians and Multi-view Generation}

As analyzed in the introduction, due to limitations in the available data, two direct approaches are not feasible:  
(i) directly generating high-quality Gaussian surfels via diffusion models;  
(ii) generating a sufficient number of multi-view images.  
Given these constraints, we aim to optimize coarse Gaussians using four-view normal and texture maps as input in the surface splatting process.

As described in the surfel representation, Gaussian surfels are characterized by the parameters \( \mathbf{p}, c, \alpha, \mathbf{s} \), and \( \mathbf{R} \). We employ a multi-head attention model to generate these Gaussian surfel parameters. First, we encode the constructed Gaussian surfels from the dataset that align with the given normal and texture latents and input them into two separate decoders: one for geometric properties (\( \mathbf{p}, \mathbf{s}, \mathbf{R} \)) and the other for texture properties (\( c, \alpha \)). Both decoders adopt multi-head attention mechanisms. The normal latent and texture latent conditions are injected into the hidden layers and jointly participate with the surfel latent in generating the final surfel parameters.

The normal latent and texture latent are obtained from two text-conditioned generative models: a Multi-view Normal Diffusion Model and a Normal-conditioned Texture Diffusion Model. The texture latent is conditioned on the encoded normal images. Furthermore, we decode both the normal and texture latents to obtain four-view normal and texture images.

During training, we fine-tune the pretrained Multi-view Normal Diffusion Model and the Normal-conditioned Texture Diffusion Model. The remaining modules are trained using a combination of mean squared error (MSE), spherical harmonic (SH) constraints, and Kullback-Leibler (KL) divergence losses:
\begin{equation}
\mathcal{L}_{T} = \mathcal{L}_{\mathrm{recon}} + \lambda_{\mathrm{sh}} \cdot \mathcal{L}_{\mathrm{sh}} + \lambda_{\mathrm{KL}} \cdot \mathcal{L}_{\mathrm{KL}},
\end{equation}
where\begin{equation}
\begin{aligned}
&\mathcal{L}_{\mathrm{KL}}=\mathrm{KL}(\mathcal{N}(\mu,\sigma^2)\parallel\mathcal{N}(0,I))\\
&\mathcal{L}_{\mathrm{sh}} = \frac{1}{BN} \sum_{b=1}^{B} \sum_{i=1}^{N} \left(1 - \frac{ \mathbf{s}_{b,i} \cdot \hat{\mathbf{s}}_{b,i} }{ \|\mathbf{s}_{b,i}\|_2 \cdot \|\hat{\mathbf{s}}_{b,i}\|_2 + \epsilon } \right),\\
&\mathcal{L}_{\mathrm{recon}} = \sum_l \lambda_l \cdot \| \mathcal{G}_l^{\mathrm{pre}} - \mathcal{G}_l^{\mathrm{target}} \|.
\end{aligned}
\end{equation}
The \( l \) in \( \mathcal{L}_{\mathrm{recon}} \) denotes the type of Gaussian parameter being reconstructed.

\subsection{Rendering Strategy}
\label{sec:Rendering}

At each optimization step, we render the object surface from four different viewpoints, producing four corresponding rendered normal maps and RGB maps. The RGB maps are compared with the ground truth texture images using the L1 norm to compute the RGB loss \( \mathcal{L}_c \), while the rendered normal maps are compared with the ground truth normal images to compute the normal loss \( \mathcal{L}_n \), also based on \( \ell_1 \) distance.

The original 2D Gaussian Splatting process relies on a large number of rendered images from multiple viewpoints, as well as an initial point cloud reconstructed via SfM. The splatting loss includes a depth distortion loss, which minimizes the distance between the camera ray and the intersection with Gaussian elements, encouraging the Gaussian weights to concentrate along the ray direction. This regularization term, denoted as \( \mathcal{L}_d \), is implemented in CUDA. Additionally, normal consistency ensures that surfaces reconstructed via 2D splatting locally approximate the actual object geometry.

The overall splatting loss is defined as:
\begin{equation}
\mathcal{L}_{\mathrm{splatting}} = \mathcal{L}_{c} + \beta \cdot \mathcal{L}_{d} + \gamma \cdot \mathcal{L}_{n}
\end{equation}

Due to our four-viewpoint optimization setting, the original splatting loss is insufficient to fully guide the reconstruction. Therefore, we introduce a surface convergence constraint and a surface curvature constraint under a 360° surround-view configuration, from which a curvature-based normal consistency is derived.

\subsubsection{Surface Curvature Constraint under 360\textdegree\ Viewpoints}

Due to the discrete spatial distribution of Gaussian surface elements, object geometry may exhibit inconsistencies across different viewpoints. To mitigate the effects of such unreliable reconstructions, we propose a 360\textdegree\ surround-view surface curvature constraint to enforce global geometric consistency. The surface normal is a vector field defined on the surface, and curvature measures the rate of change of this field, characterizing the surface geometry. For example, high curvature implies rapid changes in normals, while near-zero curvature implies small variations, indicating locally flat regions. Our key insight is that for surface regions overlapping across neighboring cameras, the curvature at a given point—expressed in world coordinates—should remain consistent.

However, the magnitude of a normal vector is not directly linearly related to the curvature. The computation of curvature involves both normal curvature and principal curvature, and the latter, along with principal directions, can be derived from normal curvature. Therefore, we only need to ensure that the normal curvature at a given point is consistent across neighboring views. To estimate normal curvature on a discrete surface, we adopt a technique based on estimating normal section curves, using the spatial distribution of neighboring points~\cite{curvature}.

For two neighboring views \( i \) and \( j \), the direct curvature consistency loss is defined as:
\begin{equation}
L = \frac{1}{|S_{\perp}|} \sum_{\mathbf{P} \in S_{\perp}} \left\| \kappa^i(\mathbf{P}) - \kappa^j(\mathbf{P}) \right\|,
\end{equation}
where \( \kappa^i(\mathbf{P}) \) denotes the estimated curvature at point \( \mathbf{P} \) from view \( i \), and \( S_{\perp} \) denotes the set of overlapping surface points from different perspectives.

Although effective, directly computing curvature at all points is computationally expensive, as it requires neighborhood-based estimations. We prove (see appendix) that enforcing normal curvature consistency is equivalent to enforcing normal vector consistency at corresponding overlapping points across views. Consequently, the final curvature-based normal consistency loss \( \mathcal{L}_{\kappa} \) becomes:
\begin{equation}
\label{Lkappa}
\mathcal{L}_{\kappa} = \frac{1}{|S_{\perp}|} \sum_{\mathbf{P} \in S_{\perp}} \left\| 1 - \hat{\mathbf{n}}_{\text{view}}(\mathbf{P}) \cdot \hat{\mathbf{n}}_{\text{near}}(\mathbf{P}) \right\|,
\end{equation}
where \( \hat{\mathbf{n}}_{\text{view}} \) and \( \hat{\mathbf{n}}_{\text{near}} \) are the unit normals rendered at point \( \mathbf{P} \) from the current and neighboring cameras, respectively. \( \mathbf{P} \) is the three-dimensional coordinate point of the overlapping area $S_{\perp}$ of the rendered surface view.

To cover the full 360\textdegree\ view, for each rendering viewpoint, we sample one or more neighboring views within a 180\textdegree\ angular range centered around the camera. However, when the overlapping region \( S_{\perp} \) is empty, the loss \( \mathcal{L}_{\kappa} \) becomes zero, reducing optimization efficiency. To address this, we use rendered depth maps to identify overlapping regions. A point \( \mathbf{p} \in S_{\perp} \) satisfies:

\begin{equation}
\begin{aligned}
&S_{\perp}^{ij} = S^i \cap S^j = \{ \mathbf{P}^i \cap \mathbf{P}^j \mid \mathbf{P}^i \in S^i,\ \mathbf{P}^j \in S^j \}, \\
&S^i = \{ \mathbf{P}^i(u,v) \mid D^i(u,v) > 0 \}, \\
&\mathbf{P}^i(u,v) = D^i(u,v) \cdot \mathbf{R}^{i\top} \mathbf{K}^{-1} (u,\ v,\ 1)^\top + \mathbf{T}^i,
\end{aligned}
\end{equation}
where \( (u, v) \) are image coordinates, \( D^i(u,v) \) is the rendered depth at pixel \( (u,v) \) from camera \( i \), \( \mathbf{K} \) is the camera intrinsic matrix, and \( \mathbf{R}^i, \mathbf{T}^i \) are the extrinsic rotation and translation matrices of camera \( i \). \( \mathbf{P}^i(u,v) \) is thus the back-projected 3D point in world space.

\subsubsection{Surface Convergence Constraints}

In our experiments, we observe that surfels may deviate from the true surface to better fit a limited number of views, which leads to redundancy and degraded rendering quality for novel viewpoints. To encourage a denser and more faithful representation of object geometry, we introduce a surface convergence constraint that promotes surfels to remain closer to the underlying surface, resulting in smoother geometry and more consistent textures.

To this end, we define the surface convergence loss \( \mathcal{L}_S \) as:
\begin{equation}
\label{LS}
\mathcal{L}_S = \frac{1}{|P^B|} \sum_{\mathbf{P} \in P^B} \| \mathbf{P} - \mathbf{c}_s \|,
\end{equation}
where \( P^B \) denotes the set of surfel points located away from the true surface, and \( \mathbf{c}_s \) is the center coordinate of the target surface.

\vspace{1mm}
\subsubsection{Final Loss}

Guided by the generation of normal and texture maps, we minimize the following overall loss function:
\begin{equation}
\mathcal{L} = \mathcal{L}_{\mathrm{splatting}} + \mathcal{L}_{\mathrm{normal}} + \lambda_{\kappa} \cdot \mathcal{L}_{\kappa} + \lambda_{s} \cdot \mathcal{L}_s,
\end{equation}

where:

- \( \mathcal{L}_{\mathrm{splatting}} \) includes the RGB texture loss \( \mathcal{L}_c \), which combines the \( \ell_1 \) norm and D-SSIM term, as proposed in~\cite{3DGaussian},

- \( \mathcal{L}_{\mathrm{normal}} \) includes the depth distortion and normal consistency regularization terms, following~\cite{2dsplatting},

- \( \mathcal{L}_{\kappa} \) is the curvature-based normal consistency loss,

- \( \mathcal{L}_s \) is the surface convergence loss, and \( \lambda_{\kappa}, \lambda_{s} \) are weighting hyperparameters.
\begin{figure}
    \centering
    \includegraphics[width=0.8\linewidth]{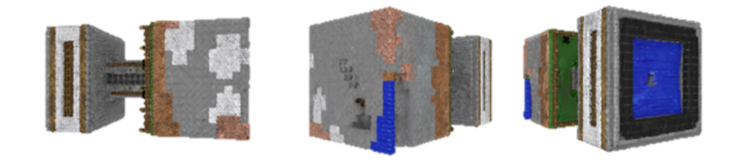}
    \caption{Caption: A floating house supported by four stone pillars above a dirt and stone block base, with a fenced grassy rooftop and a water feature at the bottom.}
    \includegraphics[width=0.8\linewidth]{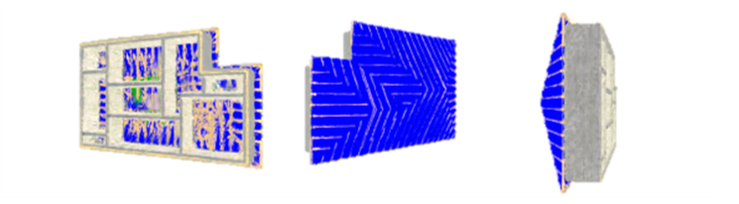}
    \caption{Caption: A rectangular building with a blue gable and hip roof structure and visible internal roof framing.}
    \label{fig:dataset}
\end{figure}
\begin{figure*}
    \centering
    \includegraphics[width=1\linewidth]{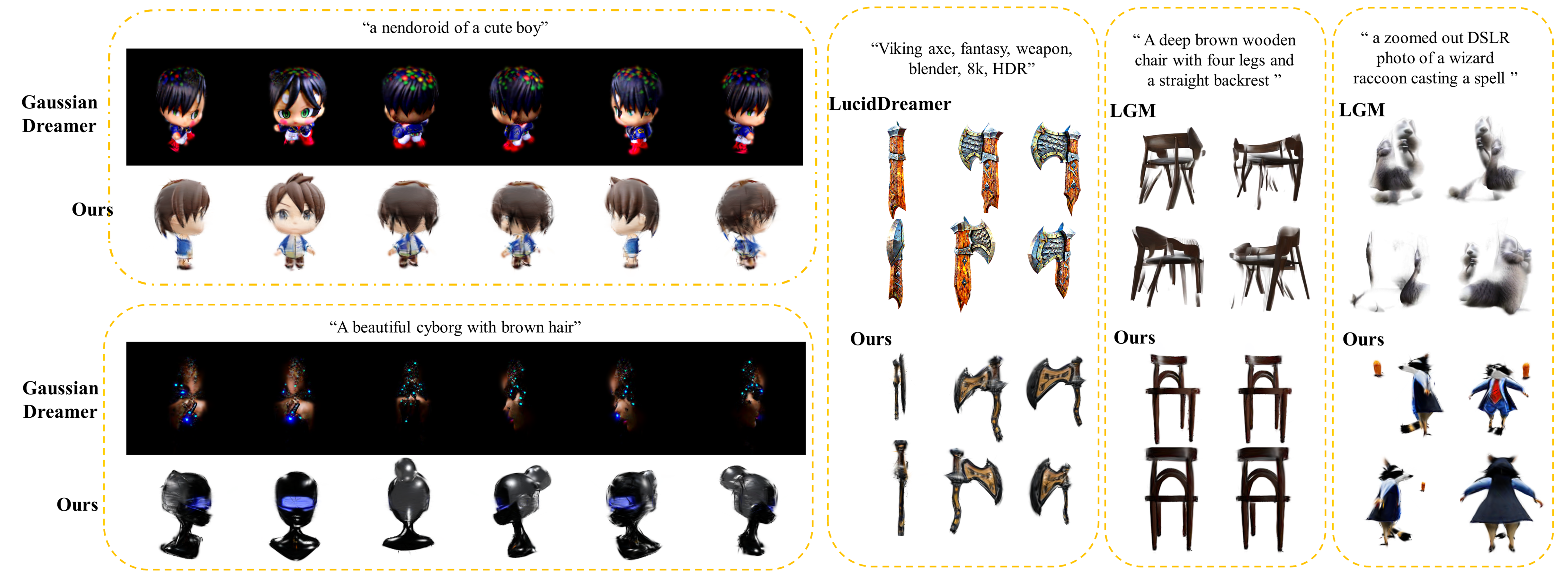}
    \caption{Comparison with three different methods based on Gaussian Splatting method. Zoom-in for details.}
    \label{fig:compare}
\end{figure*}
\section{Implementation details}
\subsection{Construction of the Real Gaussian Dataset}
As shown in the framework, training generative models of coarse Gaussian needs a set of Gaussian parameters that match the text description. We constructed the TextGaussian dataset, which maps text to Gaussian parameters, based on Cap3D and Objaverse dataset. Fig.~\ref{fig:dataset} shows two visualizations of Gaussian Serfels. Specifically, each object has a corresponding text, and we have point clouds of the object, as well as 20 rendered view images and their corresponding camera parameters. Therefore, we leverage the 2D Gaussian splatting to render a realistic Gaussian surfels representation of this object. Due to the adaptive densification strategy of Gaussian splatting, the number of surfels in TextGaussian varies across different objects, making it challenging to use the generation model with fixed parameters.
We analyzed that the target coarse Gaussian essentially does not need to be overly dense, as its role is to provide basic surfels for four-view rendering that align with the text description, effectively avoiding point cloud reconstruction and random initialization of remaining surfels parameters in 2D Gaussian splatting. Therefore, before performing the encoder in TextGaussian, we use the FPS algorithm to filter the 2D Gaussian surfels, obtaining main surfels representing the object surface and resulting in a fixed number of $1024$ surfels. Finally, the Gaussian surfels in TextGaussian are concatenated into a 58-dimensional tensor, where $\mathrm{dim}_{\mathbf{p}}=3,\mathrm{dim}_c=48,\mathrm{dim}_{\alpha}=1, \mathrm{dim}_{\mathbf{s}}=2, \mathrm{and}\  \mathrm{dim}_{\mathbf{R}}=4$.
\subsection{Model Training and Optimization}

During training, we render four views with an elevation angle of \( 0^\circ \) and azimuth angles of \( 0^\circ \), \( 90^\circ \), \( 180^\circ \), and \( 270^\circ \). The fine-tuned model is adapted from Direct2.5D\cite{direct2}, a diffusion-based architecture that generates multi-view images from text prompts. In this pipeline, the normal latent is generated first, and the texture latent is initialized using the predicted normal maps, with both conditioned on the input text.

We freeze the multi-head attention modules responsible for image generation in Direct2.5D and fine-tune two components: the Multi-view Normal Diffusion Model and the Normal-conditioned RGB Diffusion Model. Text features are encoded using the pretrained CLIP-ViT model. Gaussian surfel parameters are generated via a multi-head attention network, where the encoder and attention-based decoder are trained end-to-end.

Our experiments are conducted on NVIDIA A100 GPUs. During optimization, the surface convergence constraint is applied only during the first 2,000 iterations to allow for efficient surfel growth and surface representation. The curvature constraint is introduced after 2,500 iterations, ensuring that surfels have accumulated sufficient color and positional information beforehand. As discussed above, direct curvature computation significantly increases generation time. Therefore, we employ the indirect curvature loss defined in Equation~\ref{Lkappa} throughout our experiments.
\section{Experiment}
\begin{table}
\centering
\begin{tabular}{c@{\quad}c@{\quad}c@{\quad}c@{\quad}c}
\toprule
Method & LGM  & Direct2.5D & \textbf{Ours} \\
\midrule
CLIP Score $\uparrow$ & 32.74 & 27.56 & \textbf{32.85} \\
\bottomrule
\end{tabular}
\caption{Quantitative comparison on text-to-3D generation. For each method, the CLIP score corresponds to the best result across all text prompts. While Direct2.5D~\cite{direct2} and LDM achieve comparable CLIP scores, their high scores occur only for a few prompts.}
\label{tab:quantitative}
\end{table}
\subsection{}{Text-to-3D Gaussian Generation.}  
Figure~\ref{fig:results} presents a gallery of our generation results. We render objects from novel camera viewpoints, and the corresponding renderings align well with the text descriptions. This demonstrates the effectiveness of DirectGaussian framework under different perspectives.

\vspace{1mm}
\noindent\textbf{Qualitative Evaluation.} 
We compare our method with relevant baselines, including recent Gaussian Splatting-based Text-to-3D approaches such as \textit{GaussianDreamer}, \textit{LucidDreamer}, \textit{LGM}. Figure~\ref{fig:compare} provides a visual comparison of the rendering quality.

All methods are able to produce textured 3D objects from text inputs. Our results demonstrate improved overall consistency and authenticity. In contrast, many existing methods suffer from the Multi-face Janus problem, producing inconsistent or distorted geometry when viewed from unseen perspectives, see Fig.~\ref{fig:compare}. Our method is more robust under such conditions. Overall, our approach achieves a strong balance between generation quality and efficiency, delivering results with improved robustness.

\vspace{1mm}
\noindent\textbf{Quantitative evaluation} Our study uses 2D prior information, so we focus on measuring the consistency between text and images using the CLIP\cite{learning} metric.The CLIP score is selected as the maximum value among 240 views. Table \ref{tab:quantitative} shows the CLIP\cite{learning} results and compares the quantitative performance with methods based on image-to-3D generation. Compared with the baseline methods, our method achieves better CLIP scores and comparable generation time, demonstrating that our method excels in overall generation quality and relevance to text prompts, which are key aspects in the text-to-3D task.
\vspace{1mm}
\noindent\textbf{User Study} 
Each participant was presented with 10 different sets of rendered 3D results. Each set corresponds to a unique prompt, with results generated by 4 different methods side-by-side.
\begin{table}[htbp]
\centering
\caption{Our proposed method received the highest proportion of user preferences, indicating superior perceived quality and better alignment with text prompts in comparison to other methods.}
\begin{tabular}{lcc}
\toprule
\textbf{Method} & \textbf{Votes (\%)} & \textbf{Rank} \\
\midrule
LGM       & 10.3\% & 4 \\
GaussianDreamer           & 22.6\% & 3 \\
LucidDreamer   & 23.8\% & 2 \\
\textbf{Ours}     & \textbf{29.8\%} & \textbf{1} \\
\bottomrule
\end{tabular}
\label{tab:userstudy_results}
\end{table}

\subsection{Further Study}
\label{sec:Ablation}
\begin{figure}
    \centering
    \includegraphics[width=0.8\linewidth]{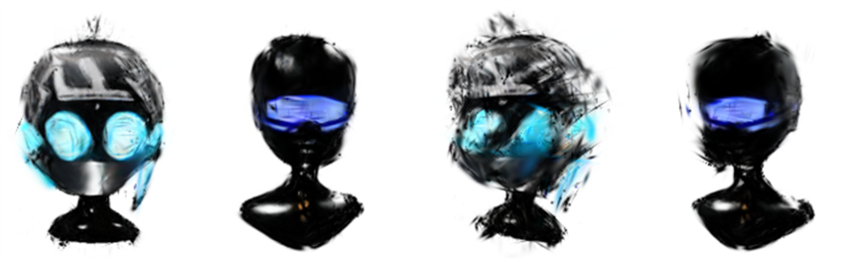}
    \caption{The caption is ``A beautiful cyborg with brown hair". Our learned coarse Gaussians lead to more stable convergence and better geometric fidelity (right). Random initialization often results in fragmented or inconsistent geometry (left).}
    \label{fig:futher1}
\end{figure}
\noindent\textbf{Effectiveness of Training Gaussians generation Model.}
In Fig.~\ref{fig:futher1}, To evaluate the importance of using the coarse Gaussian generated by our model, we compare with the framework where the initial Gaussian surfels are randomly initialized.
\begin{figure}
    \centering
    \includegraphics[width=0.5\linewidth]{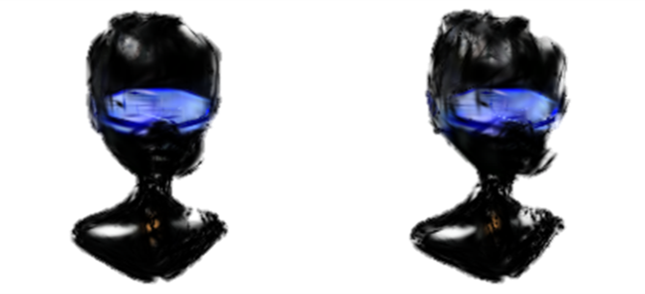}
    \caption{The caption is ``A beautiful cyborg with brown hair". Adding the curvature loss results in smoother surface transitions and improved consistency across neighboring viewpoints. Without this constraint, surfels tend to exhibit geometric noise and inconsistencies in overlapping regions.}
    \label{fig:futher2}
\end{figure}

\vspace{1mm}
\noindent\textbf{Effectiveness of Surface Curvature Constraint.}  
As shown in Fig.~\ref{fig:futher2}, We conduct a comparative study with and without the curvature constraint loss \( \mathcal{L}_\kappa \).

\section{Conclusion}

In this work, we present a novel framework for text-to-3D generation. By combining anisotropic surfels with the 2D Gaussian Splatting method, our approach can generate surfaces with realistic lighting effects and diverse geometric structures. Text input is transferred to the optimized process of coarse Gaussian through multi-view normals and texture mapping. The 360-degree curvature constraints are added to the optimization process to ensure global geometric consistency. The proposed framework opens up promising application prospects in areas such as animation, virtual reality, and video game content creation.

\bibliography{aaai2026}

\begin{thebibliography}{28}
\providecommand{\natexlab}[1]{#1}

\bibitem[{Botsch, Spernat, and Kobbelt(2004)}]{phong}
Botsch, M.; Spernat, M.; and Kobbelt, L. 2004.
\newblock Phong splatting.
\newblock In \emph{Proceedings of the First Eurographics conference on Point-Based Graphics}, 25--32.

\bibitem[{Chang et~al.(2015)Chang, Funkhouser, Guibas, Hanrahan, Huang, Li, Savarese, Savva, Song, Su et~al.}]{shapenet}
Chang, A.~X.; Funkhouser, T.; Guibas, L.; Hanrahan, P.; Huang, Q.; Li, Z.; Savarese, S.; Savva, M.; Song, S.; Su, H.; et~al. 2015.
\newblock Shapenet: An information-rich 3d model repository.
\newblock \emph{arXiv preprint arXiv:1512.03012}.

\bibitem[{Chen et~al.(2024)Chen, Li, Ye, Wang, Xie, Zhai, Wang, Liu, Bao, and Zhang}]{pgsr}
Chen, D.; Li, H.; Ye, W.; Wang, Y.; Xie, W.; Zhai, S.; Wang, N.; Liu, H.; Bao, H.; and Zhang, G. 2024.
\newblock PGSR: Planar-based Gaussian Splatting for Efficient and High-Fidelity Surface Reconstruction.
\newblock \emph{arXiv preprint arXiv:2406.06521}.

\bibitem[{Chen et~al.(2023)Chen, Chen, Jiao, and Jia}]{fantasia3d}
Chen, R.; Chen, Y.; Jiao, N.; and Jia, K. 2023.
\newblock Fantasia3d: Disentangling geometry and appearance for high-quality text-to-3d content creation.
\newblock In \emph{Proceedings of the IEEE/CVF international conference on computer vision}, 22246--22256.

\bibitem[{Dai et~al.(2024)Dai, Xu, Xie, Liu, Wang, and Xu}]{high}
Dai, P.; Xu, J.; Xie, W.; Liu, X.; Wang, H.; and Xu, W. 2024.
\newblock High-quality surface reconstruction using gaussian surfels.
\newblock In \emph{ACM SIGGRAPH 2024 Conference Papers}, 1--11.

\bibitem[{Deitke et~al.(2023)Deitke, Schwenk, Salvador, Weihs, Michel, VanderBilt, Schmidt, Ehsani, Kembhavi, and Farhadi}]{objaverse}
Deitke, M.; Schwenk, D.; Salvador, J.; Weihs, L.; Michel, O.; VanderBilt, E.; Schmidt, L.; Ehsani, K.; Kembhavi, A.; and Farhadi, A. 2023.
\newblock Objaverse: A universe of annotated 3d objects.
\newblock In \emph{Proceedings of the IEEE/CVF Conference on Computer Vision and Pattern Recognition}, 13142--13153.

\bibitem[{Huang et~al.(2024)Huang, Yu, Chen, Geiger, and Gao}]{2dsplatting}
Huang, B.; Yu, Z.; Chen, A.; Geiger, A.; and Gao, S. 2024.
\newblock 2D Gaussian Splatting for Geometrically Accurate Radiance Fields.
\newblock In \emph{ACM SIGGRAPH 2024 Conference Papers}, SIGGRAPH '24. New York, NY, USA: Association for Computing Machinery.
\newblock ISBN 9798400705250.

\bibitem[{Jung et~al.(2024)Jung, Han, An, Kang, Park, and Kim}]{rain}
Jung, J.; Han, J.; An, H.; Kang, J.; Park, S.; and Kim, S. 2024.
\newblock Relaxing Accurate Initialization Constraint for 3D Gaussian Splatting.
\newblock \emph{arXiv preprint arXiv:2403.09413}.

\bibitem[{Kerbl et~al.(2023)Kerbl, Kopanas, Leimk{\"u}hler, and Drettakis}]{3DGaussian}
Kerbl, B.; Kopanas, G.; Leimk{\"u}hler, T.; and Drettakis, G. 2023.
\newblock 3D Gaussian Splatting for Real-Time Radiance Field Rendering.
\newblock \emph{ACM Trans. Graph.}, 42(4): 139--1.

\bibitem[{Liang et~al.(2024)Liang, Yang, Lin, Li, Xu, and Chen}]{luciddreamer}
Liang, Y.; Yang, X.; Lin, J.; Li, H.; Xu, X.; and Chen, Y. 2024.
\newblock Luciddreamer: Towards high-fidelity text-to-3d generation via interval score matching.
\newblock In \emph{Proceedings of the IEEE/CVF Conference on Computer Vision and Pattern Recognition}, 6517--6526.

\bibitem[{Lin et~al.(2023)Lin, Gao, Tang, Takikawa, Zeng, Huang, Kreis, Fidler, Liu, and Lin}]{magic3d}
Lin, C.-H.; Gao, J.; Tang, L.; Takikawa, T.; Zeng, X.; Huang, X.; Kreis, K.; Fidler, S.; Liu, M.-Y.; and Lin, T.-Y. 2023.
\newblock Magic3d: High-resolution text-to-3d content creation.
\newblock In \emph{Proceedings of the IEEE/CVF Conference on Computer Vision and Pattern Recognition}, 300--309.

\bibitem[{Liu et~al.(2024)Liu, Xu, Jin, Chen, Varma~T, Xu, and Su}]{one2345}
Liu, M.; Xu, C.; Jin, H.; Chen, L.; Varma~T, M.; Xu, Z.; and Su, H. 2024.
\newblock One-2-3-45: Any single image to 3d mesh in 45 seconds without per-shape optimization.
\newblock \emph{Advances in Neural Information Processing Systems}, 36.

\bibitem[{Liu et~al.(2023)Liu, Wu, Van~Hoorick, Tokmakov, Zakharov, and Vondrick}]{zero123}
Liu, R.; Wu, R.; Van~Hoorick, B.; Tokmakov, P.; Zakharov, S.; and Vondrick, C. 2023.
\newblock Zero-1-to-3: Zero-shot one image to 3d object.
\newblock In \emph{Proceedings of the IEEE/CVF international conference on computer vision}, 9298--9309.

\bibitem[{Lu et~al.(2024)Lu, Zhang, Li, Fang, McKinnon, Tsin, Quan, Cao, and Yao}]{direct2}
Lu, Y.; Zhang, J.; Li, S.; Fang, T.; McKinnon, D.; Tsin, Y.; Quan, L.; Cao, X.; and Yao, Y. 2024.
\newblock Direct2. 5: Diverse text-to-3d generation via multi-view 2.5 d diffusion.
\newblock In \emph{Proceedings of the IEEE/CVF Conference on Computer Vision and Pattern Recognition}, 8744--8753.

\bibitem[{Mildenhall et~al.(2021)Mildenhall, Srinivasan, Tancik, Barron, Ramamoorthi, and Ng}]{nerf}
Mildenhall, B.; Srinivasan, P.~P.; Tancik, M.; Barron, J.~T.; Ramamoorthi, R.; and Ng, R. 2021.
\newblock Nerf: Representing scenes as neural radiance fields for view synthesis.
\newblock \emph{Communications of the ACM}, 65(1): 99--106.

\bibitem[{Pfister et~al.(2000)Pfister, Zwicker, Van~Baar, and Gross}]{surfels}
Pfister, H.; Zwicker, M.; Van~Baar, J.; and Gross, M. 2000.
\newblock Surfels: Surface elements as rendering primitives.
\newblock In \emph{Proceedings of the 27th annual conference on Computer graphics and interactive techniques}, 335--342.

\bibitem[{Poole et~al.(2022)Poole, Jain, Barron, and Mildenhall}]{dreamfusion}
Poole, B.; Jain, A.; Barron, J.~T.; and Mildenhall, B. 2022.
\newblock Dreamfusion: Text-to-3d using 2d diffusion.
\newblock \emph{arXiv preprint arXiv:2209.14988}.

\bibitem[{Radford et~al.(2021)Radford, Kim, Hallacy, Ramesh, Goh, Agarwal, Sastry, Askell, Mishkin, Clark et~al.}]{learning}
Radford, A.; Kim, J.~W.; Hallacy, C.; Ramesh, A.; Goh, G.; Agarwal, S.; Sastry, G.; Askell, A.; Mishkin, P.; Clark, J.; et~al. 2021.
\newblock Learning transferable visual models from natural language supervision.
\newblock In \emph{International conference on machine learning}, 8748--8763. PMLR.

\bibitem[{Rombach et~al.(2022)Rombach, Blattmann, Lorenz, Esser, and Ommer}]{stablediffusion}
Rombach, R.; Blattmann, A.; Lorenz, D.; Esser, P.; and Ommer, B. 2022.
\newblock High-resolution image synthesis with latent diffusion models.
\newblock In \emph{Proceedings of the IEEE/CVF conference on computer vision and pattern recognition}, 10684--10695.

\bibitem[{Schonberger and Frahm(2016)}]{sfm}
Schonberger, J.~L.; and Frahm, J.-M. 2016.
\newblock Structure-from-motion revisited.
\newblock In \emph{Proceedings of the IEEE conference on computer vision and pattern recognition}, 4104--4113.

\bibitem[{Shi et~al.(2023)Shi, Chen, Zhang, Liu, Xu, Wei, Chen, Zeng, and Su}]{zero123++}
Shi, R.; Chen, H.; Zhang, Z.; Liu, M.; Xu, C.; Wei, X.; Chen, L.; Zeng, C.; and Su, H. 2023.
\newblock Zero123++: a single image to consistent multi-view diffusion base model.
\newblock \emph{arXiv preprint arXiv:2310.15110}.

\bibitem[{Tang et~al.(2025)Tang, Chen, Chen, Wang, Zeng, and Liu}]{lgm}
Tang, J.; Chen, Z.; Chen, X.; Wang, T.; Zeng, G.; and Liu, Z. 2025.
\newblock Lgm: Large multi-view gaussian model for high-resolution 3d content creation.
\newblock In \emph{European Conference on Computer Vision}, 1--18. Springer.

\bibitem[{Tang et~al.(2023)Tang, Ren, Zhou, Liu, and Zeng}]{dreamgaussian}
Tang, J.; Ren, J.; Zhou, H.; Liu, Z.; and Zeng, G. 2023.
\newblock Dreamgaussian: Generative gaussian splatting for efficient 3d content creation.
\newblock \emph{arXiv preprint arXiv:2309.16653}.

\bibitem[{Wang et~al.(2023)Wang, Du, Li, Yeh, and Shakhnarovich}]{wang2023score}
Wang, H.; Du, X.; Li, J.; Yeh, R.~A.; and Shakhnarovich, G. 2023.
\newblock Score jacobian chaining: Lifting pretrained 2d diffusion models for 3d generation.
\newblock In \emph{Proceedings of the IEEE/CVF Conference on Computer Vision and Pattern Recognition}, 12619--12629.

\bibitem[{Wang et~al.(2024)Wang, Lu, Wang, Bao, Li, Su, and Zhu}]{prolificdreamer}
Wang, Z.; Lu, C.; Wang, Y.; Bao, F.; Li, C.; Su, H.; and Zhu, J. 2024.
\newblock Prolificdreamer: High-fidelity and diverse text-to-3d generation with variational score distillation.
\newblock \emph{Advances in Neural Information Processing Systems}, 36.

\bibitem[{Yi et~al.(2024)Yi, Fang, Wang, Wu, Xie, Zhang, Liu, Tian, and Wang}]{gaussiandreamer}
Yi, T.; Fang, J.; Wang, J.; Wu, G.; Xie, L.; Zhang, X.; Liu, W.; Tian, Q.; and Wang, X. 2024.
\newblock GaussianDreamer: Fast Generation from Text to 3D Gaussians by Bridging 2D and 3D Diffusion Models.
\newblock In \emph{CVPR}.

\bibitem[{Zhang et~al.(2008)Zhang, Li, Cheng et~al.}]{curvature}
Zhang, X.; Li, H.; Cheng, Z.; et~al. 2008.
\newblock Curvature estimation of 3D point cloud surfaces through the fitting of normal section curvatures.
\newblock \emph{Proceedings of ASIAGRAPH}, 2008(23-26): 2.

\bibitem[{Zwicker et~al.(2001)Zwicker, Pfister, van Baar, and Gross}]{Surfacesplatting}
Zwicker, M.; Pfister, H.; van Baar, J.; and Gross, M. 2001.
\newblock Surface splatting.
\newblock In \emph{Proceedings of the 28th Annual Conference on Computer Graphics and Interactive Techniques}, SIGGRAPH '01, 371–378. New York, NY, USA: Association for Computing Machinery.
\newblock ISBN 158113374X.

\end{thebibliography}

\end{document}